\pdfoutput=1
\documentclass[smallcondensed]{svjour3}       
\smartqed  
\def\showfigs{0}

\usepackage{natbib}
\usepackage{times}
\usepackage{verbatim}
\usepackage{subfigure}
\usepackage{amsmath}
\usepackage{gensymb}
\usepackage{natbib}
\usepackage{xfrac}
\usepackage{framed}

\usepackage{algorithm}
\usepackage{algorithmic}
\usepackage{url}
\usepackage{wrapfig}
\usepackage{graphicx} 
\usepackage{lipsum}
\ifx\showfigs\undefined
\else
\usepackage{empheq}

\usepackage{stackengine}
\fi

\usepackage{xfrac}
\usepackage{hhline}
\usepackage{color}
\usepackage[usenames,dvipsnames]{xcolor}
\usepackage{graphicx}
\newcommand{\beqa}{\begin{eqnarray}}
\newcommand{\eeqa}{\end{eqnarray}}

\newcommand{\size}[1]{\left|#1\right|}

\newcommand{\norm}[1]{\left\|#1\right\|}

\newcommand{\vx}{\vec{x}}
\newcommand{\vw}{\vec{w}}
\newcommand{\vwt}{\vw^{\mathsf{T}}}

\newcommand{\vs}{\vec{s}}

\newcommand{\DeltaTreat}{\Delta_{\text{treat}}}
\newcommand{\LambdaS}{\lambda_{\text{S}}}
\newcommand{\LambdaO}{\lambda_{\text{O}}}
\newcommand{\LambdaSOA}{\LambdaS^{\text{soa}}}
\newcommand{\LambdaAUC}{\LambdaS^{\text{auc}}}
\newcommand{\DeltaPrior}{\Delta_{\text{prior}}}
\newcommand{\DeltaPost}{\Delta_{\text{post}}}
\newcommand{\pTreat}{p_{\text{treatment response}}}

\newcommand{\synf}{\textit{SyntheticFlu}}

\newcommand{\treatProbT}{\rho_{\text{T}}}
\newcommand{\treatProbWBC}{\rho_{\text{WBC}}}
\newcommand{\treatProbTTrain}{\treatProbT^{\text{train}}}
\newcommand{\treatProbWBCTrain}{\treatProbWBC^{\text{train}}}
\newcommand{\treatProbTTest}{\treatProbT^{\text{test}}}
\newcommand{\treatProbWBCTest}{\treatProbWBC^{\text{test}}}

\renewcommand{\l}{\left}
\renewcommand{\r}{\right}

\begin{document}

\title{Learning (Predictive) Risk Scores in the Presence of Censoring due to Interventions}
\titlerunning{Learning (Predictive) Risk Scores in the Presence of Censoring due to Interventions}
\author{Kirill Dyagilev         \and
        Suchi Saria 
}

\institute{K. Dyagilev \at
        Department of Computer Science\\
        Johns Hopkins University\\
        3400 N. Charles St.\\
        Baltimore, MD 21218
        \email{kirilld@jhu.edu}           
           \and
           S. Saria \at
            Department of Computer Science\\
            Department of Health Policy \& Mgmt.\\
            Johns Hopkins University\\
            3400 N. Charles St.\\
            Baltimore, MD 21218
           \email{ssaria@cs.jhu.edu}
}

\date{Received: date / Accepted: date}

\maketitle

\begin{abstract}
A large and diverse set of measurements are regularly collected during a patient's hospital stay to monitor their health status. 
Tools for integrating these measurements into severity scores, that accurately track changes in illness severity, can improve clinicians ability to provide timely interventions.
Existing approaches for creating such scores either 1) rely on experts to fully specify the severity score, 2) infer a score using detailed models of disease progression, or
3) train a predictive score, using supervised learning, by regressing against a surrogate marker of severity such as the presence of downstream adverse events. The first approach does not extend to diseases where an accurate score cannot be elicited from experts. The second assumes that the progression of disease can be accurately modeled, limiting its application to populations with simple, well-understood disease dynamics.
The third approach, also most commonly used, often produces scores that suffer from bias due to treatment-related censoring \citep{lit:paxton_prediction_in_EHR}.
Specifically, since the downstream outcomes used for their training are observed only noisily and are influenced by treatment administration patterns,
these scores do not generalize well when treatment administration patterns change.
We propose a novel ranking based framework for disease severity score learning (DSSL). DSSL exploits the following key observation:
while it is challenging for experts to quantify the disease severity at any given time, it is often easy to compare the disease severity at two different times.
Extending existing ranking algorithms, DSSL learns a function that maps a vector of patient's measurements to a scalar severity score subject to two constraints.
First, the resulting score should be consistent with the expert's ranking of the disease severity state. Second, changes in score between consecutive periods should be smooth.
We apply DSSL to the problem of learning a sepsis severity score using a large, real-world electronic health record dataset.
The learned scores significantly outperform state-of-the-art clinical scores in ranking patient states by severity and in early detection of downstream adverse events.
We also show that the learned disease severity trajectories are consistent with clinical expectations of disease evolution.
Further, we simulate datasets containing different treatment administration patterns and show that DSSL shows better generalization performance to changes in treatment patterns compared to the above approaches.

\end{abstract}

\section{Introduction}\label{sec:intro}
\begin{figure}[h]
\centering
\includegraphics[scale = 0.30]{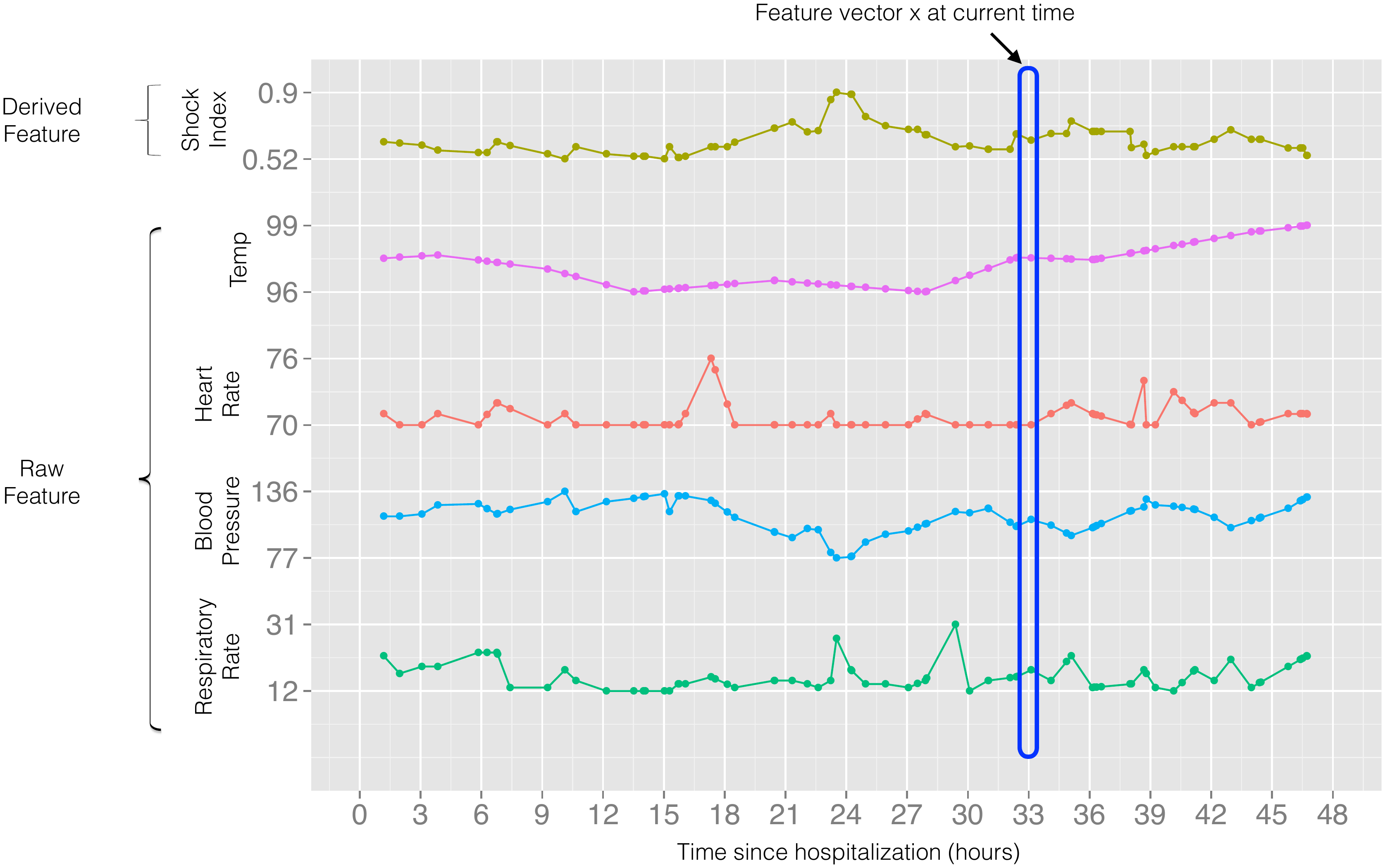}\\
\caption{Measurements over time for an example patient in an intensive care unit (ICU). In blue, we identify the feature vector $\vx^p$ at time $t=33$ hours, i.e.,
all available measurements for a patient $p$ at time $t$.}\label{fig:ex_long}
\vspace{-15pt}
\end{figure}

Consider the task of monitoring patients admitted to the Intensive Care Unit (ICU). Clinicians must regularly assess for changes in disease severity to plan timely interventions.
Since direct observation of a patient's disease state is rarely possible, assessing severity requires the caregiver to interpret a diverse array of markers
(e.g., heart rate, respiratory rate, blood counts, and serum measurements) that measure the underlying physiologic and metabolic state.
In Figure \ref{fig:ex_long}, we show a subset of such data collected on a single patient in the intensive care unit over the $48$-hour period preceding when they experienced septic shock. Continuous assessments of whether an individual is at-risk based on this data is both time-consuming and challenging.
$ $\\
In this paper, we address the problem of quantifying (scoring) the latent severity of an individual's disease at a given time.
That is, we derive a mapping from the high-dimensional observed marker data to a numeric score that tracks changes in severity of the underlying disease state over time ---
as health worsens, the score increases, and as the individual's health improves, the score declines.
Accurate estimation and tracking of the underlying disease severity can enable clinicians to detect critical decline such as decompensations, and acute adverse events in a timely manner.
Additional benefits of accurate disease severity estimation include a means for measuring an individual's response to therapy
and stratification of patients for resource management and clinical research \citep{lit:sev_scores_review_keegan}. \\
$ $\\
\textbf{Defining a Disease Severity Score:}  Qualitatively, the concept of a disease severity score has been described as the total effect of disease on the body;
the irreversible effect is referred to as damage, while the reversible component is referred to as activity \citep{lit:medsger}.
The precise interpretation of concepts of damage and activity are typically based on the application at hand. Desirable properties of a severity scale include: 1) face and content validity i.e., the variables included  are important and clinically credible, and 2) construct validity i.e., the scoring system parallels an independently ascertained severity measurement \citep{lit:medsger}. \\
$ $\\
\textbf{Prior Art:}
Historically, severity scores have been designed in a number of different ways \citep{lit:ghanem_dss_sepsis}. One approach is to have clinical experts fully specify the score.
Namely, using existing clinical literature, a panel of experts identifies factors that are most indicative of severity of the target disease.
These factors are weighted by their relative contribution to the severity and summed together to yield the total resulting score.
For example, the Acute Physiology And Chronic Health conditions score \citep{lit:apache_ii} (APACHE II), which assesses
the overall health state in an in-patient setting, uses factors that are most predictive of mortality.
A heart rate between $110$ and $139$ beats per minute adds $2$ points to the final score while a heart rate higher than $180$ beats per minute adds $4$ points. 
Similarly, mean arterial blood pressure between $70$ and $109$ mm Hg adds no points while a value between $50$ and $69$ mm Hg adds $2$ points.
A number of additional widely used scoring systems have been designed in this way, including the Multiple Organ Dysfunction Score \citep{lit:mods} (MODS),
the Sequential Organ Failure Assessment \citep{lit:sofa} (SOFA), and Medsger's scoring system \citep{lit:medsger}.

A second approach commonly taken is to assume that the severity can be characterized in terms of another surrogate measure such as the risk of an impending adverse event or mortality. This method relies on the intuition that high severity states are more likely to be associated with adverse events and higher mortality rates. The disease severity score is then learned by regressing a mapping between observed biomarkers and elements of clinical history and the risk. For instance, the pneumonia severity index (PSI) combines $19$ factors including age, vitals and laboratory test results, to calculate the probability of morbidity and mortality among patients with community acquired pneumonia \citep{lit:psi_intro}. The relative weight of each factor in the resulting score was derived by training a logistic regression predictor of patient's death in the following $30$-hour window. For simplicity of use, the relative weights were normalized so that the weight of the age would be equal to one and rounded up to the closest multiple of $10$ (of $15$ for temperature). Others have similarly used downstream adverse events such as the development of \textit{Clostridium difficile} infection \citep{lit:wiens_risk_stratification}, septic shock \citep{lit:ho_lr_for_ss}, morbidity \citep{lit:saria2010}, and mortality \citep{lit:sicula} as surrogate sources of supervision for training severity scores.

A third approach uses probabilistic state estimation techniques to track disease severity and progression (e.g., \citealp{lit:mould,lit:jackson2003multistate,lit:saria-NIPS,lit:sontag}). These model disease progression as a function of the observed measurements.
For example, \cite{lit:jackson2003multistate} study abdominal aortic aneurysms in elderly men.  They divide the progression of this disease into discrete stages of increasing severity according to successive  ranges of aortic diameter. The disease dynamics is modeled using a hidden Markov model (HMM), which allows to capture both the transition between the stages and the stage misclassification probability. The parameters of this model are estimated using maximum likelihood. Once model parameters are known, the disease severity of a patient (the unobserved state of the HMM) at a given time can be obtained by inference on the learned model.

However, all of the above-mentioned approaches for derivation of disease severity scores have their limitations. The expert-based approach captures known clinical expertise well, but does not extend to populations where the current clinical knowledge is incomplete.
The progression modeling based approaches require making assumptions about the disease dynamics and are therefore only applicable to diseases where the dynamics are relatively well-understood.
Finally, the third approach, also the most commonly used, often produces scores that suffer from bias due to treatment-related censoring \citep{lit:paxton_prediction_in_EHR}. To see why, note that for model training, supervised examples are obtained by annotating each patient's record as a positive or negative training example depending on whether they experienced the target outcome or not \citep{lit:psi_intro}. However, a high-risk patient, if treated in a timely manner, may not experience the adverse event. If there is a group of such patients, who are consistently treated and therefore never experience the adverse event, the learning algorithm will consider their symptoms preceding their treatment as low-risk states, and give it a low severity score.
This poses a problem when this severity score is moved to a different environment where treatment decisions are made based on the score alone.
A caregiver may chose not to treat these high-risk state because of their low score, thereby worsening outcomes.
We elaborate on this issue further with the $\synf$ example in Section \ref{sec:dss_for_synf}. Accounting for the effects of treatments on the downstream outcome is one way to circumvent this issue (e.g., \citealp{lit:katie_science}), in this paper we propose an alternative framework.
$ $\\$ $\\
\textbf{Our contribution.} We propose Disease Severity Score Learning (DSSL) framework that exploits this key observation that, while requesting experts to quantify disease severity at a given time is challenging,
acquiring \emph{clinical comparisons} --- clinical assessments that order the disease severity at two different times --- is often easy.
These clinical comparisons, compared to labels based on downstream adverse events, are also less sensitive to treatment patterns.
Further, in the majority of diseases, clinical guidelines provide rules for coarse-grained assessment of stages of a disease (see examples in \citealp{lit:ahrq_url}). These stages can be used to augment expert-provided clinical comparisons with those that are automatically generated using these guidelines. We show how we leverage an existing guideline \citep{lit:surviving_sepsis_campaign} in our example application.

DSSL uses clinical comparisons within the same patient and across patients to train a temporally smooth disease severity score. From these clinical comparisons, DSSL learns a function that maps the patient’s observed feature vectors to a scalar severity score. With some abuse of terminology, we refer to this mapping function as the disease severity score (DSS) function.
We present two different algorithms for learning the DSS --- the first in the linear setting, and the second in the non-linear setting. In both cases, the parameters of the DSS function are found by optimizing an objective function that contains two key terms. The first term penalizes for pairs that are incorrectly ordered by their severity. The second term imposes a penalty on changes of the severity score that are driven by the temporal evolution of the disease. For example, in our application, sepsis evolves slowly and the learning objective leverages this by penalizing scores that are not smooth over those that are.
We show how two commonly used ranking algorithms can be extended to our problem in a relatively straightforward manner. For the linear DSS, we extend the soft max-margin formulation by \cite{lit:svmrank_joachims_clickthough} to maximize separation between ordered pairs while preserving temporal smoothness. For the non-linear DSS, the score is represented non-parametrically using a weighted sum of regression trees. We use an optimization procedure similar to that of gradient boosted regression \citep{lit:boosting_as_gradient_descent,lit:gbrt_freedman} to obtain the DSS function parameters. We show numerical results on the task of training a sepsis severity score for patients in the ICU.

Below, we highlight the main strengths of the proposed DSS learning framework:
\begin{enumerate}
\item Our learning algorithm provides a scalable and automatic approach to learning disease severity scores
in new disease domains and populations.
\item Our learning algorithm only requires a means for obtaining \emph{clinical comparisons} --- ordered pairs comparing disease severity state at different times. This form of supervision is more natural to elicit than asking clinical experts to map the disease severity score,
or encoding an accurate model of disease progression. Moreover, this supervision can often be generated automatically. Our approach allows experts to tune the quality of the score by increasing the granularity and amount of supervision given.
\item We show that our algorithm learns scores that are consistent with clinical expectations. For example, changes
in the severity score over consecutive time periods are smooth and the score is higher in periods adjacent to an adverse event.
Additionally, the score is sensitive to changes in disease severity state due to therapies.
\end{enumerate}

\section{The Disease Severity Score Learning (DSSL) Framework}
In this section, we introduce our methodology for learning DSS functions.
We begin by outlining the general framework for learning a temporally smooth disease severity score in Section \ref{sec:DSS_learning}.
Section \ref{sec:mm_dss} presents the soft-margin approach for learning a linear DSS function.
In Section \ref{sec:gbrt_dss}, we extend our methodology to non-linear DSS functions using gradient boosted regression trees.

\subsection{Overview}\label{sec:DSS_learning}
We consider data that are routinely collected in a hospital setting. These include covariates such as age, gender,
and clinical history (e.g., presence or absence of a clinical condition such as AIDS or Diabetes) obtained at the time of admission;
time-varying measurements such as heart rate, respiratory rate, urine volume obtained throughout the length of stay;
and text notes summarizing the patients evolving health status.
These data are processed and transformed into tuples $<\vx^p_i, t^p_i>$ where $\vx^p_i \in \mathcal{R}^d$ is a $d$-dimensional
feature vector associated with patient $p\in P$ at time $t^p_i$ for $i \in \{1,...,T^p\}$ and $T^p$ is the total number of tuples for patient $p$.
A feature vector $\vx^p_i$ contains raw measurements (e.g., last measured heart rate or last measured white blood cell count) and features derived from one or more measurements (e.g., the mean and variance of the measured heart rate over the last six hours or the total urine output in the last six hours per kilogram of weight).
In Figure \ref{fig:ex_long} in Section \ref{sec:intro}, we showed example components of a feature vector computed for a patient in the intensive care unit over a 48 hour period.
Let $D$ denote the set of tuples across all patients in the study.

The problem of learning a DSS function is defined by the sets $O$ and $S$ of pairs of tuples from the set $D$ of all tuples,
and by the set $G$ of permissible DSS functions.
The set $O$ contains pairs of tuples $(<\vx^p_i,t^p_i>, <\vx^q_j,t^q_j>)$ that are ordered by severity based on clinical
assessments. We refer to each of these paired tuples as a \emph{clinical comparison} and the set $O$ as the set of all available clinical comparisons. For notational simplicity, we assume that $\vx_i^p$ corresponds to a more severe state than $\vx_j^q$. These clinical comparisons can be obtained by presenting clinicians with data $\vx^p_i$ for patient $p\in P$ at time $t^p_i$ and
data $\vx^q_j$ for patient $q\in P$ at time $t^q_j$.
For each such pair of feature vectors, the clinical expert identifies which of these correspond to a more severe health state;
the expert can choose not to provide a comparison for a pair where the severity ordering is ambiguous. These pairs can also be generated in an automated fashion by leveraging existing clinical guidelines. In Section \ref{sec:pairs_gen}, we describe how we use an existing guideline in our application.

The set $S$ contains pairs of tuples $(<\vx^p_i,t^p_i>, <\vx^p_{i+1},t^p_{i+1}>)$ that correspond to
feature vectors that are taken from the same patient $p$ at consecutive time steps $t^p_i$ and $t^p_{i+1}$.
These pairs are used to impose smoothness constrains on the learned severity scores. We thus refer to the pairs in $S$ as the \emph{smoothness pairs}.
Finally, the set $G$ contains a parameterized family of candidate DSS functions $g$ that map feature vectors $\vx$ to a scalar severity score.

Our goal is to identify a function $g\in G$ that quantifies the severity of the disease state
represented by a feature vector $\vx$. In particular, this function should correctly order any pair
$(\vx,\vx')$ of feature vectors by their severity, and the resulting score should be temporally smooth to
mimic the natural inertia exhibited by our biological system.
We use empirical risk minimization to identify such a function $g$.
Namely, we construct an objective function $C^g$ that maps functions $g\in G$ to their empirical risk.
The first of the two terms in $C^g$ is
\begin{equation}
\sum_{(<\vx^p_i,t^p_i>, <\vx^p_{i+1},t^p_{i+1}>)\in S}\l[\frac{g(\vx^p_{i+1})-g(\vx^p_i)}{t^p_{i+1}-t^p_i}\r]^2.\label{eq:temp_smooth_cost}
\end{equation}
This term penalizes DSS functions that exhibit large changes in the severity score over short durations,
hence encouraging selection of temporally smooth DSS functions.
The second term in $C^g$ penalizes $g$ for pairs of tuples $(<\vx^p_i,t^p_i>, <\vx^q_j,t^q_j>)\in O$
for which the severity ordering induced by $g$ on vectors $\vx^p_i$ and $\vx^q_j$ is inconsistent with the ground truth clinical assessment.
i.e., $g(\vx^p_i)<g(\vx^q_j)$. We discuss the full objective comprising these two terms in greater detail in \ref{sec:mm_dss}.

In the following two sections we describe the objectives and corresponding optimization algorithms for learning the linear and non-linear DSS functions in a new disease domain.

\subsection{Learning a Linear DSS}\label{sec:mm_dss}
We first consider the problem of learning linear DSS functions, i.e.,
DSS functions of the form $g_w(\vx) = \vwt \vx$. We refer to the corresponding learning procedure as L-DSS.

\begin{figure}
\centering
\includegraphics[scale = 0.33]{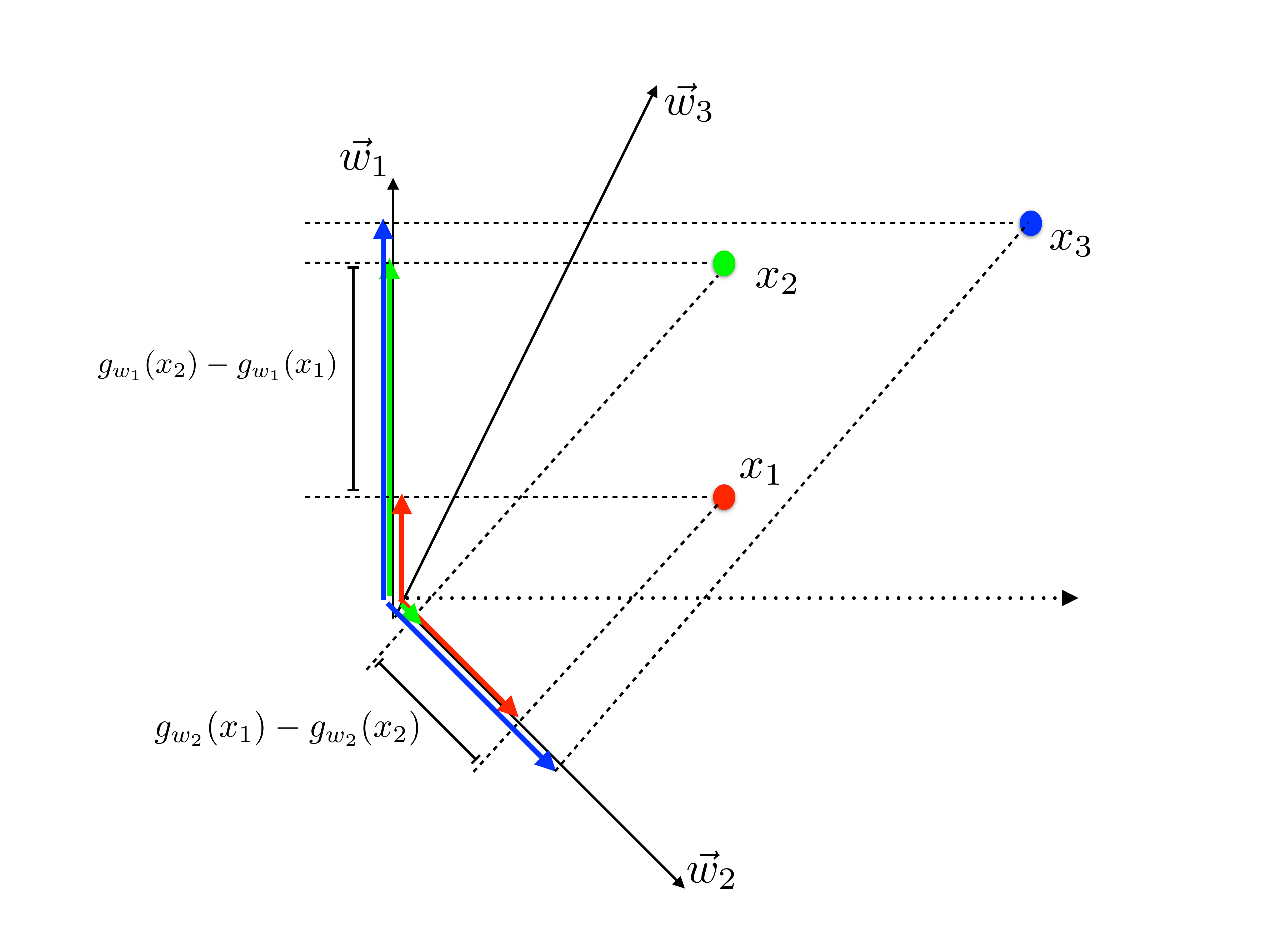}\\
\caption{We show the projections of $\vx_1$, $\vx_2$ and $\vx_3$ on vectors $\vw_1$ and $\vw_2$ representing two candidate ranking functions; vectors drawn in red, green and blue identify projections of
$\vx_1$, $\vx_2$ and $\vx_3$ respectively. Ranking is induced by the differences in the projections---for example, $\vw_2$ induces the ordering $g_{\vw_2}(\vx_1) > g_{\vw_2}(\vx_2)$ because $g_{\vw_2}(\vx_1)-g_{\vw_2}(\vx_2) > 0$.}\label{fig:margin}
\vspace{-15pt}
\end{figure}

We employ soft max-margin training \citep{lit:svmrank_joachims_clickthough} where we seek to maximize the distance
between the pairs that are at different severity levels while keeping the distance between the consecutive pairs
smooth. We briefly review the key concepts of soft max-margin ranking before we describe our extension for learning a linear DSS function given data.\\
$ $\\
\textbf{Soft Max-Margin Ranking:}
Consider the toy example shown in Figure \ref{fig:margin}. Let $D$ contain the three feature vectors $\{\vx_1, \vx_2,\vx_3\}$ where $\vx_i \in \mathcal{R}^2$, and $O$ contain the pairs $(\vx_2,\vx_1)$ and $(\vx_3,\vx_2)$,
i.e., feature vectors $\vx_2$ and $\vx_3$ have higher disease severity than $\vx_1$ and $\vx_2$ respectively.
Max-margin ranking seeks to find a vector $\vw$ such that the \emph{margin} between pairs of different severity levels is maximized. In our example, we show parameter vectors $\vw_1$, $\vw_2$ and $\vw_3$ for three candidate ranking functions in Figure \ref{fig:margin}. For each feature vector $\vx$, the assigned (severity) score for a given ranking function parameter $\vw_i$ is computed as the projection, $g_{\vw_i}(\vx)$, of $\vx$ on $\vw_i$. The induced ranking between two vectors $\vx_1$ and $\vx_2$ is computed based on the margin which is defined as the difference in their projections.
In the example shown, the rankings induced by both $g_{\vw_1}$ and $g_{\vw_3}$ correctly order all pairs in $O$, i.e.,
\begin{center}
$g_{\vw_1}(\vx_3) > g_{\vw_1}(\vx_2) > g_{\vw_1}(\vx_1) \text{ and } g_{\vw_3}(\vx_3) > g_{\vw_3}(\vx_2) > g_{\vw_3}(\vx_1)$, \\
\end{center}
while the rankings induced by $\vw_2$ do not. Furthermore, $\vw_3$ also induces an ordering with a larger margin between the pairs in $O$. Margin-maximization leads to an ordering that is more robust with respect to noise in $\vx$.

More formally, for each pair of feature vectors $(\vx_i,\vx_j)\in O$,
we define the margin of their separation by the function $g_{\vw}(\cdot)$ as $\mu_{i,j}^{\vw} = g_{\vw}(\vx_i)- g_{\vw}(\vx_j)$. The maximum-margin approach
suggests that we can improve generalization and robustness of the learned separator by selecting $\vw$ that
maximizes the number of tuples that are ordered correctly (i.e., $\mu_{i,j}^{\vw}>0$) while simultaneously
maximizing the minimal normalized margin $\mu_{i,j}^{\vw}/\norm{\vw}$.
Using the standard soft max-margin framework, the SVMRank algorithm \citep{lit:svmrank_joachims_clickthough} approximates the above-mentioned problem as the
following convex optimization program:
\begin{eqnarray}
&&\min_{\vw,\space \zeta^{i,j}_{\text{O}}}\l[\frac{1}{2}\norm{\vw}^2 +
\frac{\LambdaO}{\size{O}}\sum_{(\vx_i, \vx_j)\in O}\zeta^{i,j}_{\text{O}}\r]\label{eq:mm_opt}\\
&&\text{subject to the following }\textbf{ordering constraints}:\notag\\
&&\forall (\vx_i, \vx_j)\in O:\quad g_{\vw}(\vx_i) - g_{\vw}(\vx_j) \geq 1-\zeta^{i,j}_{\text{O}}\text{ and }\zeta^{i,j}_{\text{O}}\geq 0\notag
\end{eqnarray}

$ $\\
\textbf{The L-DSS Objective and Optimization Algorithm:}
We now describe our algorithm for learning linear DSS functions. We return to our original setting where we are given sets $O$ and $S$ which contain feature vectors belong to more than one patients at varying times.

We augment the soft-max margin objective with the additional term, shown in Eq. \eqref{eq:temp_smooth_cost}, that encourages temporal smoothness. We state the full L-DSS objective below.
\begin{eqnarray}
&&\min_{\vw,\space \zeta^{ij}_{\text{O}}}\l[\frac{1}{2}\norm{\vw}^2 +
\frac{\LambdaO}{\size{O}}\sum_{(<\vx^p_i,t^p_i>, <\vx^q_j,t^q_j>)\in O}\zeta^{(p,i), (q,j)}_{\text{O}}\r.\label{eq:mm_opt}\\
&&\phantom{\min_{\vw,\space \zeta^{(p,i), (q,j)}_{\text{O}}}\quad} + \l.\frac{\LambdaS}{\size{S}}\sum_{(<\vx^p_i,t^p_i>, <\vx^p_{i+1},t^p_{i+1}>)\in S}\l[\frac{g_{\vw}(\vx^p_{i+1})-g_{\vw}(\vx^p_i)}{t^p_{i+1}-t^p_i}\r]^2\r]\notag\\
&&\text{subject to the following }\textbf{ordering constraints}\text{:}\notag\\
&&\forall (<\vx^p_i,t^p_i>, <\vx^q_j,t^q_j>)\in O:\quad g_{\vw}(\vx^p_i) - g_{\vw}(\vx^q_j) \geq 1-\zeta^{(p,i), (q,j)}_{\text{O}}\quad\text{ and }\notag\\
&&\forall (<\vx^p_i,t^p_i>, <\vx^q_j,t^q_j>)\in O:\quad \zeta^{(p,i), (q,j)}_{\text{O}}\geq 0\notag
\end{eqnarray}
Here, the coefficients $\LambdaO$ and $\LambdaS$ control the relative degree of emphasis on the smoothness versus the margin-maximization component of the objective. For a given setting of $\LambdaO$, different choices of $\LambdaS$ yield trajectories with differing levels of smoothness. An appropriate choice of $\LambdaS$ could be determined by the clinical user based on the rate of change in severity that is to be expected in that domain. For example, in sepsis, changes in severity do not occur within minutes while in many cardiac conditions, rapid changes in severity can occur. Alternately, this parameter can be set using cross-validation to optimize performance for a particular application of DSS.

In Eq. \eqref{eq:mm_opt}, for every value of $\vw$, the optimal values of $\zeta^{(p,i), (q,j)}_{\text{O}}$ are given by
\begin{eqnarray}
\zeta^{(p,i), (q,j)}_{\text{O}} &=& \max\{0, 1 - (g_{\vw}(\vx^p_i)-g_{\vw}(\vx^q_j))\}.\label{eq:slack_hinge}
\end{eqnarray}
Substituting Eq. \eqref{eq:slack_hinge} and $g_{\vw}(\vx)=\vwt\vx$ in Eq. \eqref{eq:mm_opt}, we obtain the following unconstrained convex optimization formulation:
\begin{eqnarray}
&&\min_{\vw}\quad \frac{1}{2}\norm{\vw}^2 + \frac{\LambdaO}{\size{O}}\sum_{(<\vx^p_i,t^p_i>, <\vx^q_j,t^q_j>)\in O}\max\{0, 1 - \vwt(\vx^p_i-\vx^q_j)\} \label{eq:ldss_hinge_loss_formulation}\\
&&\phantom{\min_{w,\space \zeta^{ij}_{\text{O}}}\quad} + \frac{\LambdaS}{\size{S}}\sum_{(<\vx^p_i,t^p_i>, <\vx^p_{i+1},t^p_{i+1}>)\in S}\l[\frac{\vwt(\vx^p_{i+1}-\vx^p_i)}{t^p_{i+1}-t^p_i}\r]^2\notag
\end{eqnarray}
Instead of solving the dual formulation as in \cite{lit:svmrank_joachims_clickthough}, following the reasons of
efficiency and accuracy discussed by \cite{lit:chapelle_efficient_ranking}, we solve the primal form of this optimization program as follows.

The terms of the form $\max\{0,a\}$, also called the hinge loss, are not differentiable at $a=0$. We approximate these
terms with the Huber loss $L_h$ for $0<h<1$ given by
\begin{eqnarray*}
L_h(a) = \begin{cases}
  0&, \text{ if }a<-h\\
  \frac{(a + h)^2}{4h}&, \text{ if }\size{a}\leq h\\
  a&, \text{ if }a>h
  \end{cases}
\end{eqnarray*}
This approximation yields the following unconstrained, convex, twice-differentiable optimization problem:
\begin{empheq}[innerbox=\fbox,left=\def\stackalignment{l}\stackanchor{L-DSS}{Objective}:\space]{align}
&&\min_{\vw}\quad \frac{1}{2}\norm{\vw}^2\phantom{+ \frac{\LambdaO}{\size{O}}\sum_{(<\vx^p_i,t^p_i>, <\vx^q_j,t^q_j>)\in O}L_h(1)aaaaaaaaa}\notag\\
&&+ \frac{\LambdaO}{\size{O}}\sum_{(<\vx^p_i,t^p_i>, <\vx^q_j,t^q_j>)\in O}L_h(1 - \vwt(\vx^p_i-\vx^q_j))\phantom{\min_{\vw,\space}}\label{eq:ldss_huber_loss_formulation}\\
&& + \frac{\LambdaS}{\size{S}}\sum_{(<\vx^p_i,t^p_i>, <\vx^p_{i+1},t^p_{i+1}>)\in S}\l[\frac{\vwt(\vx^p_{i+1}-\vx^p_i)}{t^p_{i+1}-t^p_i}\r]^2\notag
\end{empheq}

We solve this optimization program using the Newton-Raphson algorithm.
We show experiments using the L-DSS learner in Section \ref{sec:exp}.

\subsection{Learning a Non-linear DSS}\label{sec:gbrt_dss}
In many disease domains, assuming a linear mapping between the measurements and the latent disease severity may be too restrictive. For example, ranges for measurements values that are considered to be \emph{normal} (or from a low-severity state) are often age dependent or clinical history dependent. Consider an individual with a pre-existing kidney condition; he or she is likely to have a worse baseline creatinine level (a test that measures kidney function) compared to an individual with fully-functioning kidneys. Thus, when measuring changes in severity related to the kidney, these individuals are likely to manifest a disease differently. See the guideline by \cite{lit:surviving_sepsis_campaign} for other examples.

To learn non-linear DSS functions, we represent $g$ as a weighted sum of regression trees. Alternate choices for learning non-linear DSS functions exist including extending the soft-margin formulation presented for learning L-DSS via
use of the ``kernel-trick" \citep{lit:kernel_svm}. We chose to extend boosted regression trees as this is one of the most widely used algorithms for ranking (e.g., see \citealp{lit:gbrt_mohan2011web}).

Our hypothesis class $G$ includes all linear combinations of shallow regression trees, i.e., functions of the form
$ g(\vx) = \sum_{k = 1}^{K} \alpha_{k} f_{k}(\vx), $
where $f_{k}$ for $k=1,...,K$ are shallow (limited-depth) regression trees and $K$ is finite. In our experiments, K is set to 5.
Similar to the objective for L-DSS in Eq. \eqref{eq:ldss_huber_loss_formulation}, we construct the NL-DSS objective to identify $g \in G$ that maximizes the dual criteria of ordering accuracy and temporal smoothness as:

\begin{empheq}[innerbox=\fbox,left=\def\stackalignment{l}\stackanchor{NL-DSS}{Objective}:\space]{align}
&&C^g(g) = \frac{1}{\size{O}}\sum_{(<\vx^p_i,t^p_i>, <\vx^q_j,t^q_j>)\in O}L_h(1 - (g(\vx^p_i)-g(\vx^q_j)))\label{eq:C_g}\\
&&+ \frac{\LambdaS}{\size{S}}\sum_{(<\vx^p_i,t^p_i>, <\vx^p_{i+1},t^p_{i+1}>)\in S}\l[\frac{g(\vx^p_{i+1})-g(\vx^p_i)}{t^p_{i+1}-t^p_i}\r]^2\notag
\end{empheq}
Note that since the soft max-margin formulation is not defined for a non-linear classifier we drop the term $\norm{\vw}^2/2$. Thus, without loss of generality, $\LambdaO$ can be replaced by $1$. Now, the relative emphasis on the smoothing versus the ordering components are changed by varying $\LambdaS$.

We optimize the NL-DSS objective using the gradient boosted regression trees (GBRT) learning algorithm \citep{lit:gbrt_freedman,lit:boosting_as_gradient_descent,lit:lambda_mart}. Gradient boosting methods grow $g$ incrementally, in a greedy fashion, by adding a weak learner---in this case, a regression tree---at each iteration. A tree that most closely approximates the gradient of $C^g$ evaluated at $g$ obtained in the previous iteration is added \citep{lit:gbrt_freedman}.

The per-iteration computational complexity of this approach is equivalent to the computational complexity of building a single regression tree, which is $\size{T}\log\size{T}$  \citep{lit:rpart}, where $\size{T}$ is the number of unique tuples in the set $O\cup S$ of tuple pairs.

\section{Experiments}\label{sec:exp}
We now describe the evaluation of the proposed DSSL framework. Before discussing numerical results on a real-world dataset, in Section \ref{sec:dss_for_synf}, we use a simple toy example to illustrate
the behavior of DSS related to the following questions. First, when DSS is transported between environments with different degrees of interventional confounds,
how is the performance of DSS impacted compared to the performance of a supervised learning algorithm that uses downstream-events as labels?
Second, when clinical comparisons are generated by implementing automated coarse-grading rules, does the learned score simply learn the rule itself?
In Section \ref{sec:data_desc}, we provide background on our application: we introduce sepsis, the dataset used, and the guideline used for generating the clinical comparisons needed to train the DSS scores. Next, in Section \ref{sec:dss_mimic}, we provide an overview of the experiments and the experimental setup followed by a detailed discussion of the numerical results on the sepsis data in Section \ref{sec:exp_results}.

\subsection{Learning DSS for $\synf$}\label{sec:dss_for_synf}
For these experiments, we create a simple toy disease called $\synf$ as follows. We quantify severity as a function of the patient's temperature and white blood cell counts (WBC)---as the temperature or the WBC increases, risk of mortality increases. We assume that the disease manifests in two ways: with $50\%$ probability, patients are sampled from a model where the temperature tends to \emph{deteriorate} while the WBC remains \emph{normal}, and for the other fraction of the population, their WBC tends to deteriorate while the temperature remains normal. Each of these measurements assume one of $10$ states (e.g., the temperature ranges from $99$ to $108$ \degree F). In the absence of treatment, for a measurement that is deteriorating, it retains its value $T$ in the following timestep with probability $0.3$, increases to $T+1$ with probability $0.5$, and decreases to $T-1$ with probability $0.2$. For a measurement that is assumed to stay normal, the corresponding transition probabilities are $0.7$, $0.1$ and $0.2$.
States $1-2$ are defined to be ``benign''(e.g., temperatures of 99\degree F and 100\degree F) where an individual can be discharged with probability $0.5$ (i.e. their sampled trajectory ends).
States $6-9$ are defined to be ``severe'' (e.g., temperatures between 104\degree F to 107\degree F) where an individual may receive treatment with probability $\treatProbT$. Administration of treatment (e.g., antibiotics) transitions the individual to one of the benign states. Finally, an individual dies when she or he reaches state $10$. \\
$ $\\
\textbf{Evaluating transportability:}
The first question we investigate is regarding the transportability of the different scores. Namely, when DSS is moved between different treatment regimes, how
is the performance of DSS impacted compared to the performance of a supervised learning algorithm that uses downstream-events as labels. Since treatments affect the prevalence of adverse outcomes, we show that the risk scores learned via the latter approach are highly sensitive to treatment patterns and therefore, in our example setting, generalize poorly compared to DSS. We sample $1000$ patients each for the train and test sets. Data are sampled from different treatments regimes as shown in Table \ref{tab:toyExampleResults}.
For example, in scenario $1$, no treatments are prescribed in either the train or test regimes.
In scenario $5$, in the train regime, treatments are prescribed with $0.3$ probability only for treating high temperature but not for high WBC. However, in the test regime, treatments are prescribed only for high WBC but not for high temperature. Training L-DSS and NL-DSS requires generating clinical comparisons. To do so, we randomly sample pairs from the observed data. For a pair $(\vx_i^{p}, \vx_j^{q})$, we consider $\vx_i^p$ to represent a more severe state if one of the measurements is at least 2 units higher in $\vx_i^p$ than in $x_j^q$ (e.g., temperature of 103\degree F is more severe than 101\degree F)  and the other measurement is at least as high in $\vx_i^p$ as in $\vx_j^q$.
Many such pairs are sampled for the train and test set. We use a standard protocol for training logistic regression (LR) \citep{lit:paxton_prediction_in_EHR}. Train and test samples are generated from the patient trajectories via a sliding window approach with the outcome defined as whether or not the patient died within $10$ timesteps in the future. On the test set, we consider a patient to be correctly identified as at-risk patient if his severity score was greater than a certain threshold value at any point of patient's hospital stay.
We measure performance of the obtained scores using the area under the curve (AUC) obtained on the task of predicting per-patient mortality \citep{lit:paxton_prediction_in_EHR}.\\
$ $\\
Results: In scenario $1$ and scenario $2$ where the treatment patterns are the same across the train and test regimes, all three scores perform equally well. However, as the treatment patterns begin to diverge to an increasing degree as seen in scenarios $4$ and $5$, LR's performance drops while the DSS performance does not change.
It is worth noting that such discrepancies between the train and test regimes can occur within the same hospital when comparing treatment practice before and after deploying a predictive model. Specifically, while clinicians continue to treat high-risk states, the resulting LR score learned from this data will underestimate risks for the treated states. Once the decision support system is deployed and clinicians begin to rely on the predictive tool, they may erroneously undertreat high-risk patients, thereby worsening outcomes.
\begin{table}[ht]
\centering
\begin{tabular}{||l||c|c|c|c||c|c|c||}
\hline
Scenario & $\treatProbTTrain$ & $\treatProbWBCTrain$ & $\treatProbTTest$ &  $\treatProbWBCTest$ & Logistic Regression & L-DSS & NL-DSS\\
\hline\hline
\#1 & 0 & 0 & 0 & 0 & 0.974 & 0.973 & 0.974\\
\#2 & 0.1 & 0 & 0.1 & 0 & 0.978 & 0.990 & 0.991\\
\#3 & 0.1 & 0 & 0 & 0 & 0.963 & 0.974 & 0.981\\
\#4 & 0.3 & 0 & 0 & 0 & 0.769 & 0.973 & 0.981\\
\#5 & 0.3 & 0 & 0 & 0.3 & 0.510 & 0.978 & 0.996\\
\hline
\end{tabular}
\caption{Analysis of transportability of the L-DSS, NL-DSS and logistic regression based severity scores between different treatment regimes. $\treatProbTTrain$, $\treatProbWBCTrain$, $\treatProbTTest$, and $\treatProbWBCTest$ denote the probability of treatment for temperature and WBC in the train and test regimes respectively.}\label{tab:toyExampleResults}
\vspace{-20pt}
\end{table}
$ $
\\
$ $ \\
\textbf{Relationship of Learned Scores to the Coarse Grades:} As mentioned, for many diseases, clinical guidelines provide rules for coarse-grained assessment of severity stages of a disease.
These guidelines can be used for automated generation of clinical comparison pairs with umabiguous severity ordering.
This raises a natural question of whether a DSS learned from such clinical comparisons simply recovers these clinical guidelines and thus yields no generalization beyond the coarse grading.
To evaluate this hypothesis, we extend the setup described above. Not sure if the previous sentence gives too strong of a claim. After all, we show a single example where this does not happen.
Specifically, we augment the feature vectors to include the coarse grades which are in turn derived from the temperature and WBC measurements.
We derive the coarse severity grade from the observed feature vectors as follows. We assign a feature vector $\vx_i^p$ a severity of $0$ if both the corresponding WBC and temperature measurements are in states $1$ or $2$ (e.g., temperature below 101\degree F), a severity of $1$ if exactly one of the measurements is in state $3$ or higher (e.g., temperature above or equal to $101$\degree F), and a severity of $2$ if both measurements are in states $3$ or higher. We consider all three combinations of coarse graded severity pairs and randomly sample $6000$ clinical comparisons and a similar number of smoothness pairs. We use $\LambdaO = 100$ for the L-DSS and sweep values $\LambdaS$ between $0.1$ and $1000$. We show results using data sampled from regime $1$ though the conclusions do not depend on the treatment pattern.

\begin{figure}[h]
\centering
\includegraphics[scale = 0.4]{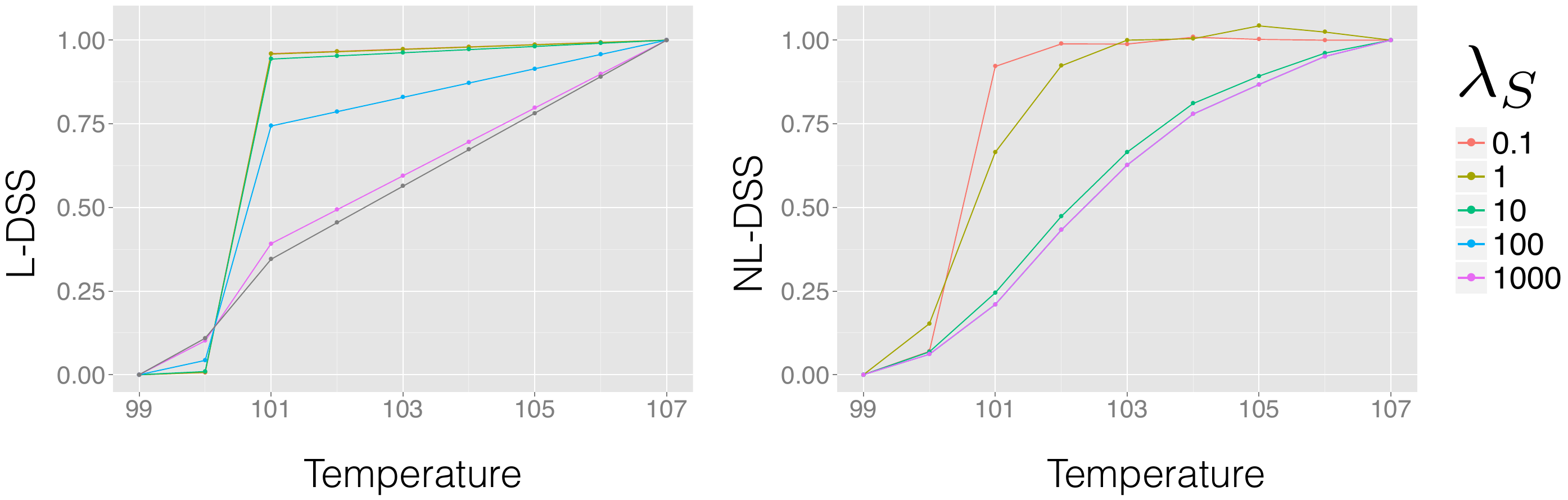}\\
\caption{Experiments on synthetic data: relationship of learned scores to the coarse grades. L-DSS and NL-DSS scores for varying values of $\LambdaS$ for a patient with WBC in state 2 and temperature ranging
from $99$\degree F to $107$\degree F. The scores are normalized so that temperature of $99$\degree F is given the score of $0$ and the temperature of $107$ \degree F is given the score of $1$.}\label{fig:synth_prob}
\vspace{-25pt}
\end{figure}

Since perfect ordering accuracy can be achieved using only the coarse grading component of the feature vector, one might expect that the learned scores will rely on the coarse grading feature alone.
In fact, this is the case when $\LambdaS$ is small ($=0.1$).
However, such DSS score will exhibit abrupt changes between consecutive time points, e.g.,  when the temperature or WBC progresses from state 2 to state 3 or vice versa.
As $\LambdaS$ increases, the smoothness term in the DSS objective encourages the learning of temporally smooth scores which rely increasingly on the WBC and temperature features alone.
Thus, in this scenario, the smoothness constraint allows the L-DSS and NL-DSS scores to generalize beyond coarse grades.
In Figure \ref{fig:synth_prob}, we depict the severity score assigned to a feature vector in which WBC is in state 1 and temperature varies from $99$\degree F to $107$\degree F
for varying values of $\LambdaS$.

\subsection{Sepsis, MIMIC-II and the Surviving Sepsis Campaign Guideline}\label{sec:data_desc}
In the following experiments on the real-world clinical data, our goal is to learn a score that assesses the severity of sepsis.\\
$ $\\
\textbf{Sepsis:} Sepsis is a whole-body inflammatory response to infection; it is a leading cause of death in the inpatient setting, with especially high mortality among patients who develop septic shock, a major sepsis-related adverse event. Both sepsis and septic shock are known to be associated with high morbidity, longer hospital stay and increased health care cost \citep{lit:kumar_ss_trends}. Often, the risk of sepsis-related adverse outcomes can be reduced by early treatment \citep{lit:effects_of_early_response}, thus a scoring system that allows precise tracking of changes in sepsis-related disease severity is of great importance.\\
$ $\\
\textbf{Dataset:} We use MIMIC-II, a publicly available dataset containing electronic health record data from patients admitted
to the ICUs at the Beth Israel Deaconess Medical Center from $2001$ to $2008$ \citep{lit:MIMICII_ref_1, lit:MIMICII_ref_2}. We only include adults ($>15$ years old) in our study ($N=16,234$). We compute $45$ different features that derived from vital sign measurements, clinical history variables, and laboratory test results. We provide the complete list of features in Appendix \ref{app:feat_desc}. We impute missing data using linear interpolation.
Other examples of imputation methods used for ICU monitoring datasets include model-based (e.g., \citealp{lit:ho_lr_for_ss}) and Last Observation Carried Forward (LOCF) \citep{lit:hug_thesis}.
These methods require making numerous domain-specific assumptions. For example, in LOCF, each measurement is carried forward for a finite time window which length is determined based on the typical sampling frequency of this measurement.
Since this choice is orthogonal to the focus of our paper, we implement linear interpolation as it is the simplest of the above methods.\\
$ $\\
\textbf{Guideline for grading sepsis severity:}
In order to train a severity score in the DSSL framework, we need a set $O$ of pairs of feature vectors ordered by their severity. We create these pairs automatically by leveraging the coarse severity grading of sepsis established in the Surviving Sepsis Campaign Guideline (SSCG) \citep{lit:surviving_sepsis_campaign}. The SSCG provides rules for identifying when an individual is in each of the four stages of severity: septic shock, severe sepsis, SIRS and none. For each of these stages, the guideline defines criteria using 1) a combination of thresholds for individual measurements, and 2) presence of specific diagnosis codes or diagnoses noted in their clinical notes. For example, the SIRS criterion is met when values of at least two out of these five features are out of their normal range. White blood cell count per microliter, for example, is considered to be out of the normal range if its value is either below $4.0$ or above $12.0$. Heart rate is considered to be out of the normal range if it is above $90$ beats-per-minute. The stage of severe sepsis is reached when physician suspects that patient developed an infection, the SIRS criterion is met, and at least one organ system is showing signs of failure. Finally, septic shock is defined as severe sepsis with observed hypotension despite significant fluid resuscitation. In Table \ref{tab:sscg_definitions}, we specify the criteria used for grading each of the stages. It is also worth noting that the guideline does not provide a grade at all times: on average, application of these criteria grades less than $40\%$ of data entries of a patient. We use the time points with available SSSG grading to generate clinical comparison as described in Section \ref{sec:pairs_gen}.
We also note that the SSSG grades use features beyond those used in the severity score.
For example, to assess the severity grade, we use keyword search in transcripts written at the time of discharge to determine whether the patient had developed an infection.
Similarly, features such as whether or not the patient received sufficient fluids are used for grading but excluded when learning the severity scores as these are caregiver driven and only indirectly measure severity.

\subsection{Learning DSS for sepsis in ICU patients}\label{sec:dss_mimic}
We begin with an overview of the experiments.
In our first experiment, we assess the quality of the trained L-DSS and NL-DSS scores
by their performance on the task of distinguishing between the different severity stages of sepsis.
This is done by calculating severity ordering of held out pairs of feature vectors and measuring their
concordance with the ground truth provided by the SSCG guideline.
We show that our scores significantly outperform the three main scoring systems that are widely used in ICUs.
\begin{table}[ht]
\centering
\begin{tabular}{l|l|}
\hline
Stage& Criteria\\
\hline
SIRS & At least two out of the following four conditions hold:\\
 & 1. Heart rate is $>90$ beats-per-minute and was measured in the last 2 hours.\\
 & 2. Temperature is either $>38\celsius$ or $<36\celsius$ and was measured in the last 8 hours.\\
 & 3. Respiratory rate is $>20$ beats-per-minute and was measured in the last 2 hours, or \\
 & \phantom{3. }arterial partial pressure of $CO_2$ is $<32$ mm Hg and was measured in the last 8 hours.\\
  & 4. White blood cell count in thousands per microliter is either $>12.0$ or $<4.0$ and\\
  & \phantom{4. }was measured in the last 8 hours.\\
\hline
Severe & Patient's clinical record contain words sepsis or septic or ICD-9 code for infection, \\
Sepsis & SIRS criteria holds, and at least one of the following nine criteria holds:\\
& 1. Systolic blood pressure is $<90$ mm Hg and was measured in the last 2 hours.\\
& 2. Blood lactate measurement is $>2.0$ micromol per liter and was taken in the last 2 hours.\\
& 3. Urine output over the past two hours is $<0.5$ milliliter per kg.\\
& 4. Patient has no chronic renal insufficiency and her or his creatinine measurement is $>2.0$\\
& \phantom{4.} milligrams per deciliter, and was taken in the last 8 hours.\\
& 5. Patient has no chronic liver disease, her or his bilirubin measurement is $>2.0$\\
& \phantom{4.} milligrams per deciliter and was taken in the last 8 hours.\\
& 6. Platelet count is $<100.000$ per microliter and was measured in the last 8 hours.\\
& 7. International normalized ratio (INR) is $>1.5$ and was measured in the last 8 hours.\\
& 8. Patient experienced pneumonia during their hospital stay as indicated by ICD-9 codes, \\
& \phantom{8.}measurements of both partial arterial pressure of oxygen ($PaO_2$) and of fraction of inspired\\
& \phantom{8.}oxygen ($FiO_2$) were taken in the last 8 hours, and it holds that $PaO_2/FiO_2<200$.\\
& 9. Patient experienced acute lung infection unrelated to pneumonia during their hospital stay\\
& \phantom{9.} as indicated by ICD-9 codes, the measurements of $PaO_2$ and $FiO_2$ were taken\\
& \phantom{9.} in the last 8 hours, and it holds that $PaO_2/FiO_2<250$.\\
\hline
Septic & The following two conditions hold:\\
Shock & 1. The patient has severe sepsis.\\
& 2. The patient experiences hypotension (i.e., systolic blood pressure $<90$ mm Hg)\\
& \phantom{2.} for at least last 30 minutes. \\
\hline
None & The following two conditions hold:\\
& 1. Heart rate was measured in the last two hours, temperature was measured in the last\\
& \phantom{1.}   $8$ hours, respiratory rate was measured in the last 2 hours, arterial partial pressure\\
& \phantom{1.}   of $CO_2$ was measured in the last 8 hours, and white blood cell count was measured\\
& \phantom{1.}in the last 8 hours.\\
& 2. At most one of the SIRS conditions holds. \\
\hline\hline
\end{tabular}
\caption{Definition of the Surviving Sepsis Campaign Guideline (SSCG) for sepsis severity grading.}\label{tab:sscg_definitions}
\vspace{-20pt}
\end{table}
The success of our scores in distinguishing between sepsis stages is encouraging, but expected, since our scores are explicitly trained for severity ordering. In the following experiments, we evaluate whether the learned scores also generalize well to measuring fine-grained changes in severity. Towards this, we first examine whether DSS is sensitive to changes in severity state leading up to adverse events. We consider septic shock, an adverse event of sepsis, and measure whether the learned severity scores increase leading upto septic shock as one would expect. Indeed, we show that the L-DSS and NL-DSS scores show a significant upward trend in the time period leading up to the adverse event.

Next, we evaluate whether scores trained by L-DSS and NL-DSS are sensitive to changes in severity state due to therapy.
Specifically, we compare the trend of the disease severity score before and after \emph{fluid bolus}, a therapy used to relieve
hypotension in septic patient \citep{lit:surviving_sepsis_campaign}. We show that the learned scores show significant change in their trend around the time of treatment administration, thus indicating sensitivity to treatment responses. For instance, a DSS score that trends upward over the time period leading up to administration of fluid bolus, is likely to trend down during the time period after treatment administration or to trend up at a slower pace.

Motivated by the results showing sensitivity to impending adverse events, in the last experiment,
we measure the performance of the learned severity scores for early detection of septic shock. We train
a simple classifier using the DSS and its trend features to predict risk of septic shock onset in the next $48$ hours.
We show that this predictor significantly outperforms routinely used clinical scoring systems. 

The rest of this section proceeds as follows.
In Section \ref{sec:pairs_gen} we describe our experimental setup and the procedure for the automated generation of clinical comparison pairs. Section \ref{sec:baselines} presents the baseline methods.
Finally, in Section \ref{sec:exp_results} we present the numerical results and analysis.

\subsubsection{Experimental Setup and Automatic Generation of the Clinical Comparison Pairs}\label{sec:pairs_gen}
We begin by randomly dividing the $16,234$ patients in our dataset into training ($60\%$) and testing sets ($40\%$).
Within the training set, we assign two thirds of the patients to the development set and the remaining third to the validation set. For each of the development, validation and testing sets of patients we generate a separate set $O$ of clinical comparison pairs.
We consider six combinations of possible pairs of different sepsis stages, i.e.,
none-SIRS, none-severe, none-shock, SIRS-severe, SIRS-shock, severe-shock.
For each combination of stages, we randomly select an equal number of feature vectors $(\vx_i^p, \vx_j^q)$
sampled at time points $(t_i^p, t_j^p)$ such that $\vx_i^p$ corresponds to a more severe state and $\vx_i^p$ corresponds to a less severe state in this combination. For the development and testing sets we sample $2000$ clinical comparisons for each combinations of sepsis severity stages resulting in total of $12000$ clinical comparisons for each set. For the validation set that contains only half of the number of patients in the development and testing sets, we sample $1000$ clinical comparisons for each combination of sepsis severity stages resulting in total of $6000$ clinical comparisons.

\subsubsection{Baselines: Routinely Used Clinical Severity Scores in the ICU}\label{sec:baselines}
We compare the performance of the learned disease severity scores to
three widely used ICU-based severity scoring systems \citep{lit:sev_scores_review_keegan}.
The first two scores are based on the Sequential Organ Failure Assessment or the SOFA score \citep{lit:sofa} which was originally designed to assess sepsis-related organ damage severity. The SOFA method scores severity at the per-organ level.
Two variants of the SOFA that are commonly used are: 1) total SOFA computed as the sum of SOFA scores of all organ systems,
and 2) worst SOFA represented as the highest value of SOFA score among of all organ systems.
We also compare the performance of our score to  that of the Acute Physiology and Chronic Health Evaluation or APACHE II \citep{lit:apache_ii},
which is a widely used scoring system for assessing general (not necessarily sepsis-related) disease severity in hospitalized individuals.

\subsubsection{Performance Evaluation of Scores Trained Using the L-DSS and NL-DSS Algorithms}\label{sec:exp_results}
In this section, we present numerical results of evaluation of the learned L-DSS and NL-DSS scores.\\
$ $\\
\textbf{Selection of free parameters.}
The L-DSS and NL-DSS algorithms contain free parameters $\LambdaO$ and $\LambdaS$ that remain to be specified.
Let us first consider the L-DSS algorithm. 
With  $\LambdaS$ set to $0$, we sweep the values of $\LambdaO$ from $1$ to $10^{15}$ and
select the value of $\LambdaO$ that maximizes accuracy of ordering held out pairs on the validation set.
That is, we count the fraction of ordering pairs in the set $O$ that are concordant with the ordering prescribed by the ground truth comparisons. We refer to this quantity as the severity ordering accuracy or SOA.
In the evaluations below, $\LambdaO$ is thus set to $10^{7}$.

\begin{figure}[ht]
\centering
\includegraphics[scale = 0.4]{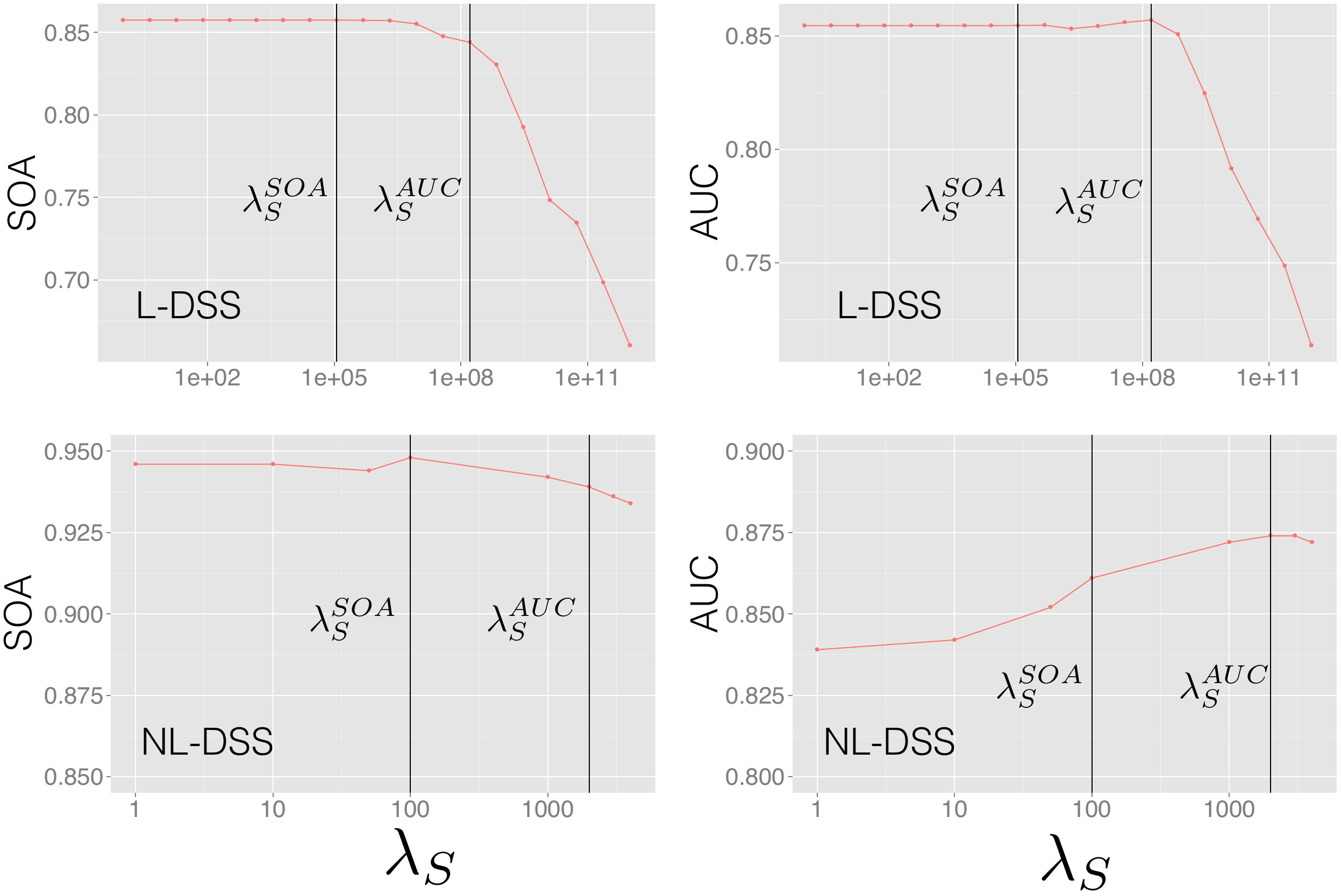}\\
\caption{Experiment 1. Severity ordering accuracy and the shock prediction AUC on the validation set for different values of $\LambdaS$. We mark the values of $\LambdaS$ selected for further evaluations, i.e., $\LambdaSOA$ and $\LambdaAUC$, with vertical lines.}\label{fig:ldss_soa_auc}
\vspace{-20pt}
\end{figure}

For a given $\LambdaO$, different choices of $\LambdaS$ yield trajectories with differing levels of smoothness.
An appropriate choice of $\LambdaS$ can be made by sweeping through a wide range of values and selecting the value that optimizes performance for a use case that the end user has in mind. For example, if the primary application of the score is for the early detection of individuals at risk for septic shock, the $\LambdaS$ that maximizes prediction accuracy on the validation set is selected. Alternately, when no assumptions are given, $\LambdaS$ can be selected to maximize smoothness without hurting ordering accuracy on the validations set. We present results using these two approaches for setting $\LambdaS$; we refer to these settings as $\LambdaAUC$ and $\LambdaSOA$ respectively. In Figure \ref{fig:ldss_soa_auc}, we show performance on the validation set and mark the selected setting of $\LambdaS$ for each of the scores. Thus, $\LambdaAUC$ and $\LambdaSOA$ were set to be $1.62\cdot 10^8$ and $1.13\cdot 10^5$ for L-DSS and $2000$ and $100$ for NL-DSS. 
While we do not experiment with this approach, it is worth noting that yet another means for selecting $\LambdaS$ is based on an expert's knowledge of the degree of short-term to long-term variability expected within that disease domain. For example, in slowly evolving diseases, the value of $\LambdaS$ that yields a small ratio of short-term to long-term variability may be preferable.

$ $\\
\textbf{Experiment 1: Distinguishing between the severity stages of sepsis. }
We begin by evaluating whether L-DSS and NL-DSS can distinguish and correctly order the different stages of sepsis severity.
We compare their severity ordering accuracy to that of routinely used clinical scores --- APACHE-II, Total SOFA and Worst SOFA \citep{lit:sev_scores_review_keegan}.
The results of this evaluation are presented in Table \ref{tab:pair_ord_prec}.
\begin{table}[ht]
\centering
\begin{tabular}{l|l|cc}
\hline
&Method & $\LambdaS = \LambdaSOA$ & $\LambdaS = \LambdaAUC$\\
\hline
Proposed &L-DSS & 0.860 & 0.844\\
Scores &NL-DSS & 0.946 & 0.938\\
\hline\hline
Routine & APACHE II & \multicolumn{2}{|c}{0.68} \\
Clinical & Total SOFA & \multicolumn{2}{|c}{0.63} \\
Scores & Worst SOFA & \multicolumn{2}{|c}{0.63} \\
\hline\hline
\end{tabular}
\caption{Experiment 1. Severity ordering accuracy (SOA) for different methods. The 95\% confidence interval on SOA is obtained using the bootstrap algorithm
and is contained in the $\pm 0.002$ band around SOA.}\label{tab:pair_ord_prec}
\vspace{-20pt}
\end{table}

We observe that L-DSS and NL-DSS significantly outperform APACHE II, Total SOFA and Worst SOFA scores for all considered values of $\LambdaS$.
The performance achieved by L-DSS and NL-DSS is significant from a clinical standpoint as it orders severity states more accurately than the three clinical scores, all of which are widely used to assess severity of ICU patients.
In particular, the SOFA score was designed specifically to measure sepsis related severity.

While the above-mentioned result is promising, it remains to be seen whether the obtained scores are also sensitive enough to capture changes in severity status that extend
beyond the coarse grading that the different stage definitions provide.
Towards this, we next evaluate whether the learned scores exhibit the following desirable characteristics:
1) Are they sensitive to changes in severity leading up to septic shock, an adverse event of sepsis? and,
2) Are they sensitive to post-therapy changes in severity?\\
$ $\\
\textbf{Experiment 2: Are the learned DSS sensitive to changes in severity leading up adverse events?}
To address the question of whether the learned scores are sensitive enough to capture changes in severity that can occur
leading up to an adverse event, we examine the L-DSS and NL-DSS behavior in the $18$ hour duration leading up to septic shock.

We consider all patients with septic shock in our test set with at least $18$ hours of data prior to septic shock onset ($N=587$). On these patients, we define three time intervals of interest: 1) $6$ hours prior to the onset of septic shock; 2) $6-12$ hours prior to the onset of the septic shock; 3) $12-18$ hours prior to the onset of septic shock.
We denote the average values of the learned scores in these intervals by $\overline{s}_{0-6}$, $\overline{s}_{6-12}$, and $\overline{s}_{12-18}$, respectively.

\begin{figure}[ht]
\centering
\includegraphics[scale = 0.4]{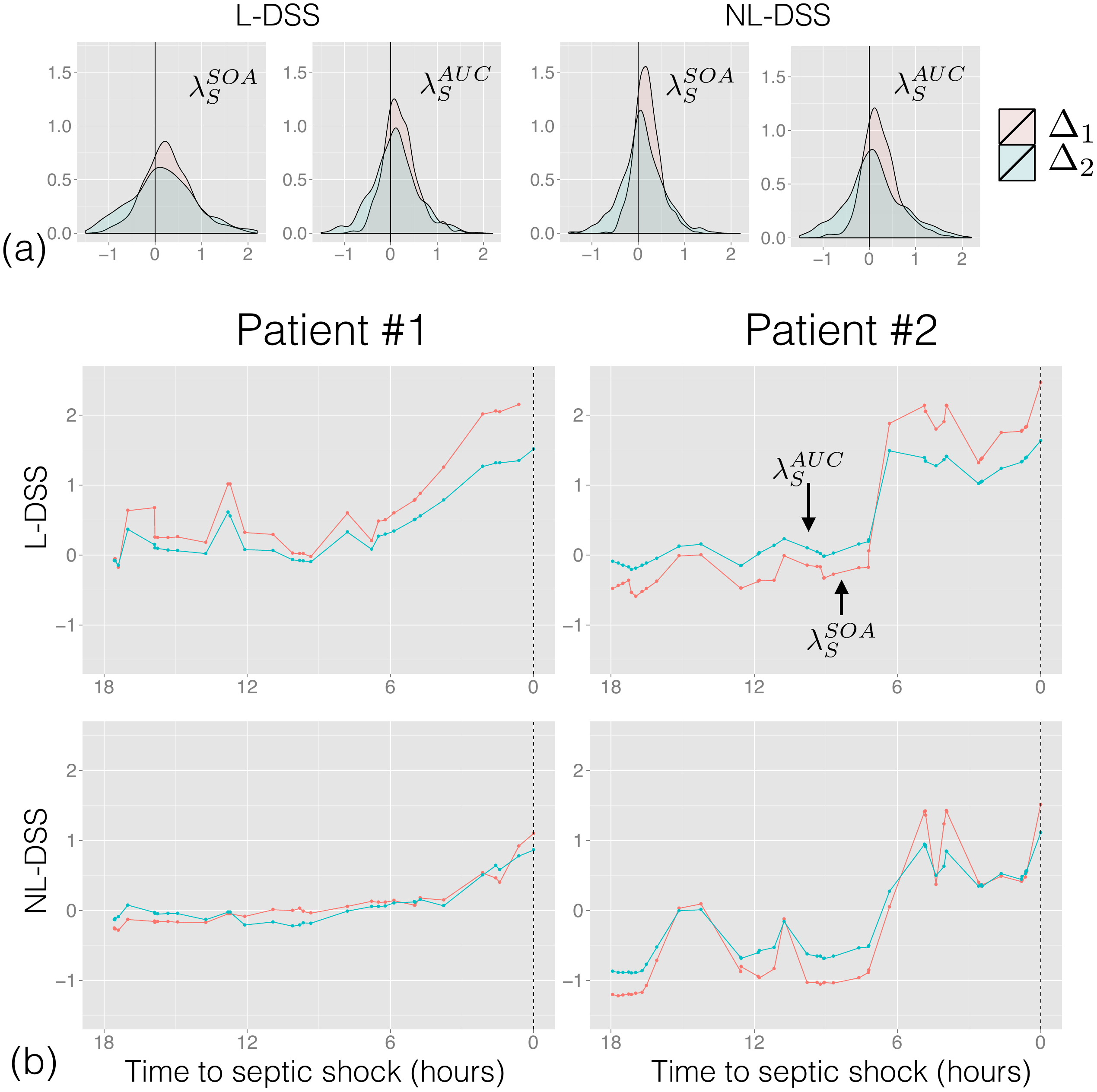}\\
\caption{Experiment 2. Sensitivity of learned DSS to changes in severity leading up to adverse events. (a) Probability density of 
$\Delta_1$ and $\Delta_2$; (b) DSS trajectories over the 18 hour period leading up to septic shock for two example patients.}\label{fig:trendup}
\vspace{-15pt}
\end{figure}

We calculate values of $\Delta_1 = \overline{s}_{0-6}-\overline{s}_{6-12}$ and $\Delta_2 = (\overline{s}_{0-6}-\overline{s}_{6-12})-(\overline{s}_{6-12}-\overline{s}_{12-18})$ for each patient. In Figure \ref{fig:trendup} (a) we show the full probability density of $\Delta_1$ and $\Delta_2$. The value of $\Delta_1$ is positive in at least $70\%$ of the cases for all four considered scores. The value of $\Delta_2$ is positive in at least $57\%$ of the cases. Using the standard one-tailed t-test, we assess the p-value (denoted by $p_{\text{trend-up}}$ in Table \ref{tab:trendup}) for whether the recorded $\Delta_1$ can be observed by chance under the null hypothesis that $\Delta_1$ are drawn from a zero mean distribution.  Similarly, we assess the p-value (denoted by $p_{\text{rate acceleration}}$ in Table \ref{tab:trendup}) for whether the recorded $\Delta_2$ can be observed by chance under the null hypothesis that $\Delta_2$ are drawn from a zero mean distribution. Across all values of $\LambdaS$, for both the L-DSS and the NL-DSS, the obtained p-values rule out the null hypothesis, that is, the learned scores leading upto septic shock show significant upward trend and acceleration. Using the bootstrap, we estimate the median p-value for a range of sample sizes and significance is achieved (i.e., the median p-value for that sample size is below 0.01) with as few as $30$ samples for $\Delta_1$ and $420$ for $\Delta_2$. 
As an example, in Figure \ref{fig:trendup}(b), we show the L-DSS and NL-DSS trajectories for two patients for the period leading up to septic shock.
\begin{table}[ht]
\centering
\begin{tabular}{l|lcc}
\hline
Method &  $\LambdaS = \LambdaSOA$ & $\LambdaS = \LambdaAUC$\\
\hline
&\multicolumn{2}{|c}{$p_{\text{trend-up}}$}\\
\hline
L-DSS & $ 10^{-36}$  &$ 10^{-39}$\\
NL-DSS & $10^{-39}$  & $10^{-46}$ \\
\hline\hline
&\multicolumn{2}{|c}{$p_{\text{rate increase}}$}\\
\hline
L-DSS & $1.3\cdot 10^{-5}$  & $4\cdot 10^{-4}$ \\
NL-DSS & $1.3\cdot 10^{-3}$  & $2.6\cdot 10^{-3}$\\
\hline\hline
&\multicolumn{2}{|c}{fraction of positive $\Delta_1$ (95\% confidence interval)}\\
\hline
L-DSS & $0.71$ (0.68-0.75)  & $0.70$ (0.67-0.74)\\
NL-DSS & $0.74$ (0.70-0.77) & $0.75$ (0.71-0.78)\\
\hline\hline
& \multicolumn{2}{|c}{fraction of positive $\Delta_2$ (95\% confidence interval)}\\
\hline
L-DSS & $0.58$ (0.54-0.62)  & $0.57$ (0.53-0.61)\\
NL-DSS & $0.59$ (0.56-0.63) & $0.58$ (0.55-0.63)\\
\hline\hline
\end{tabular}
\caption{Experiment 2. p-value $p_{\text{trend-up}}$ for the observed $\Delta_1$; p-value $p_{\text{rate acceleration}}$ for the observed $\Delta_2$;
the fraction of positive $\Delta_1$ with 95\% confidence interval, and
the fraction of positive $\Delta_2$ with 95\% confidence interval.
Calculations of all values are based on $587$ examples.}\label{tab:trendup}
\vspace{-15pt}
\end{table}

$ $\\
\textbf{Experiment 3: Are the learned DSS sensitive to post-therapy changes in severity?}
We now evaluate whether the scores trained by the L-DSS and NL-DSS methods are sensitive to changes in severity state due to administration of fluid bolus-- a treatment used for septic shock \citep{lit:surviving_sepsis_campaign}.
Towards this, we use the self-controlled case series method. We compare trends exhibited by DSS values over the five hour intervals prior to and post the administration of fluid bolus. 
We refer to the trends over these intervals as $\DeltaPrior$ and $\DeltaPost$.
The value of $\DeltaPrior$ is computed as the difference between the value of the DSS at the time of treatment administration and
the mean value of DSS over the five hour interval prior to treatment administration. 
Similarly, the value of $\DeltaPost$ is calculated as the difference between the mean value of DSS over the five hour interval after treatment administration and 
the value of DSS at the moment of treatment administration. If the patient is responsive to fluid therapy, then $\DeltaTreat = \DeltaPost - \DeltaPrior <0$, that is, 
if the DSS was trending up prior to treatment administration, we expect this trend to be attenuated or even reversed by the treatment.

We identify cases of fluid administration events related to sepsis using the following criteria:
1) the patient is experiencing SIRS, severe sepsis or septic shock at the time of treatment administration, and
2) the patient is hypotensive (has systolic blood pressure below $100$ mm Hg), a commonly used criteria for prescribing fluids in sepsis.
To avoid confounding due to multiple administration of fluids, we restrict our attention to treatment administrations that were not preceded
or followed by another fluid bolus administration within a five hour window. This yielded a total of $81$ fluid bolus administration events.

\begin{figure}[ht]
\centering
\includegraphics[scale = 0.45]{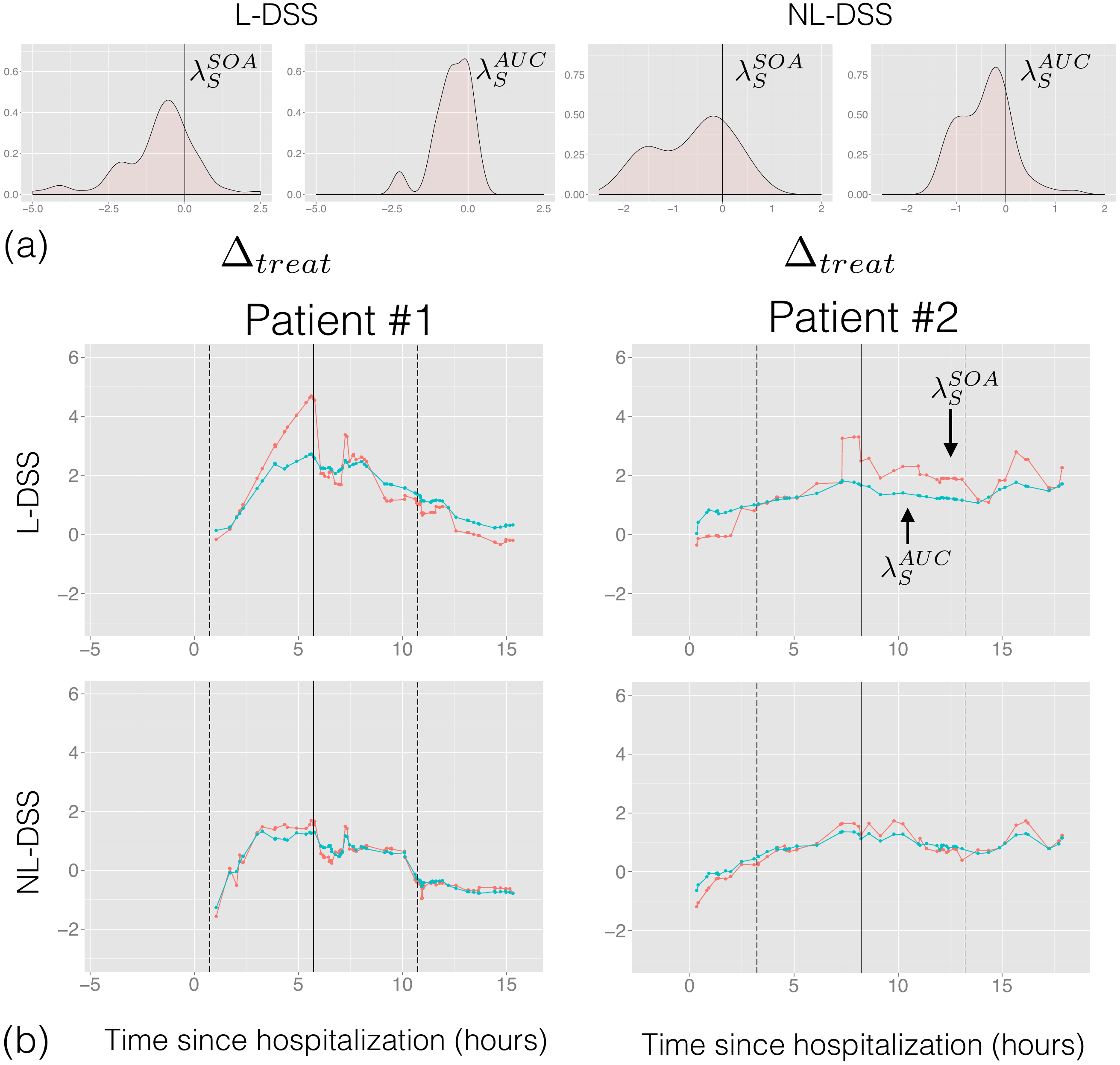}\\
\caption{Experiment 3. Sensitivity of DSS to post-treatment changes in severity. (a) Probability density of $\DeltaTreat$; 
(b) DSS trajectories before and after administration of fluid bolus for two example patients.}\label{fig:treatResponse}
\vspace{-15pt}
\end{figure}

In Figure \ref{fig:treatResponse} (a) we plot the distribution of $\DeltaTreat=\DeltaPost-\DeltaPrior$. 
Overall, the change of trend $\DeltaTreat$ is negative in at least $75\%$ of recorded values of $\DeltaTreat$.  
Employing the one-tailed t-test, we obtain the p-value $\pTreat$ (shown in Table \ref{tab:treatResponse}) for whether the observed values 
of $\DeltaTreat = \DeltaPost-\DeltaPrior$ can be observed by chance under the null hypothesis that $\DeltaTreat$ are drawn from a zero mean
distribution. For our sample size of $81$ cases, across all values of  $\LambdaS$ for both the L-DSS and NL-DSS, 
the obtained p-values rule out the null hypothesis in favor of the stated hypothesis, that is, DSS shows significant response to therapy. Moreover, using the boostrap to estimate the median p-value for a range of samples sizes, we observe that significance is achieved with as few as $20$ samples. In Figure \ref{fig:treatResponse} (b), we show the L-DSS and NL-DSS trajectories for two example patients around the time of fluid bolus administration.
\begin{table}[ht]
\vspace{-15pt}
\centering
\begin{tabular}{l|lcc}
\hline
Method &  $\LambdaS = \LambdaSOA$ & $\LambdaS = \LambdaAUC$\\
\hline
&\multicolumn{2}{|c}{$\pTreat$} \\
\hline
L-DSS & $5\cdot 10^{-7}$  & $2\cdot 10^{-10}$\\
NL-DSS & $5.1\cdot 10^{-8}$ &  $4.8\cdot 10^{-10}$\\
\hline\hline
&\multicolumn{2}{|c}{fraction of negative $\DeltaTreat$ (95\% confidence interval)} \\
\hline
L-DSS & $0.84$ ($0.77$--$0.93$)  & $0.77$ ($0.68$--$0.85$)\\
NL-DSS & $0.75$ ($0.67$--$0.85$) & $0.84$ ($0.77$--$0.93$)\\
\hline\hline
\end{tabular}
\caption{Experiment 3. Statistical significance of DSS response to fluid bolus treatment. p-value $\pTreat$ for the observed $\DeltaTreat$;
the fraction of negative $\DeltaTreat$ with 95\% confidence interval. Calculations of all values are based on $81$ examples.}\label{tab:treatResponse}
\vspace{-15pt}
\end{table}
\\\textbf{Predictive Score for Septic Shock Using DSS:} The high ordering performance in experiment $1$ and the significant upward trend observed in experiment $2$ suggest that a score such as DSS maybe useful for detecting individuals at risk for septic shock. Thus, we conclude this section by showing that the value of the severity score and its temporal trajectory can be used in prediction tasks, specifically, in the task of early detection of septic shock. 

We begin by describing the trend features that we derive from sequences of instantaneous values of severity scores.
For a patient $p$, we let $s_i^p$ for $i=1,...,T_p$ be the value of the assigned severity scores at time $t_i^p$.
At every time point $t_i^p$, we augment the score value $s_i^p$ with the seven derived trend features that were inspired by
\cite{lit:wiens_risk_stratification} and adapted to the specifics of the MIMIC II dataset.
We present the complete list of these features in Table \ref{tab:derived_feat}. In this table,
Feature $1$ is the average value of the score since admission. We note that the feature vectors
are not sampled at a fixed rate, i.e., the length of the time interval between two consecutive
feature vectors of a patient need not be fixed. We thus weigh every value of the severity score
with the length of the time interval between the current feature vector and the previous one.
Features $2$ and $3$ are versions of Feature $1$ where more recent scores have a higher relative weight.
Feature $4$ captures the average rate of change in severity score since admission.
Feature $5$ is a version of Feature $4$ in which more recent rates of score change are given higher relative weight.
Finally, Features $6$ and $7$ capture the variability of the score.
\begin{table}[ht]
\centering
\begin{tabular}{cll}
\hline\noalign{\smallskip}
Feature & Description & Expression\\
\noalign{\smallskip}\hline\hline\noalign{\smallskip}
1 & Average score since admission & $\l(1/t^p_i\r)\sum_{j=1}^i (t^p_{j}-t^p_{j-1})s_{j}^p$\\
\noalign{\smallskip}\hline\noalign{\smallskip}
2 & Linear weighted score & $\sum_{j=1}^i \l(t_{j}^p\cdot s_{j}^p\r)/\l(\sum_{j=1}^i t_{j}^p\r)$\\
\noalign{\smallskip}\hline\noalign{\smallskip}
3 & Quadratic weighted score & $\sum_{j=1}^i \l((t_{j}^p)^2 s_{j}^p\r)/\l(\sum_{j=1}^i (t_{j}^p)^2\r)$\\
\noalign{\smallskip}\hline\noalign{\smallskip}
4 & Average score change rate & $\l(s_{i}^p-s_{1}^p\r)/\l(t_{i}^p-t_{1}^p\r)$\\
\noalign{\smallskip}\hline\noalign{\smallskip}
5 & Linearly weighted average score change rate & $\sum_{j=1}^i(s_{j}^p-s_{j-1}^p)\cdot\frac{t_{j}^p-t_{j-1}^p}{t_{i}^p-t_{1}^p}$\\
\noalign{\smallskip}\hline\noalign{\smallskip}
6 & Average absolute score change rate & $\sum_{j=1}^i\size{s_{j}^p-s_{j-1}^p}\cdot\frac{t_{i}^p-t_{1}^p}{t_{i}^p-t_{1}^p}$\\
\noalign{\smallskip}\hline\noalign{\smallskip}
7 & Linearly weighted absolute score change rate & $\sum_{j=1}^i\size{s_{j}^p-s_{j-1}^p}\cdot\frac{t_{j}^p-t_{j-1}^p}{t_{i}^p-t_{1}^p}$\\
\noalign{\smallskip}\hline
\end{tabular}
\caption{Trend features derived from the trajectory of the severity score. }\label{tab:derived_feat}
\vspace{-20pt}
\end{table}

To train a predictive score for the task of early identification of sepsis, we use logistic regression to learn a mapping from a patient's feature vectors to the probability of occurrence of septic shock in the following $48$ hours. This approach was inspired by \citep{lit:LR_for_mort_pred,lit:minne,lit:ho_lr_for_ss}. Specifically, we use logistic regression regularized with an elastic net penalty \citep{lit:elastic_net}. As positive training examples, 
we consider feature vectors $\vx_i^p$ taken less than $48$ hours prior to an adverse event. As negative training examples, we take feature vectors from patients that do not experience the adverse event during their hospital stay and do not receive any treatment of fluid-bolus. The choice of leaving out individuals who receive fluid-bolus but do not experience septic shock is owing to the fact that their outcome is censored due to treatment \citep{lit:paxton_prediction_in_EHR}. We refer to this procedure as \emph{LR-Shock}. Employing LR-Shock with severity score and its trend features as its input, we obtain a predictor of the onset of septic shock in the next $48$ hours. We refer to predictors based on the L-DSS and NL-DSS scores and their derived features as LR-Shock+L-DSS+Derived and LR-Shock+NL-DSS+Derived, respectively. As a baseline for comparison, we train LR-Shock+$\vx$, which is a predictor trained by LR-Shock with feature vectors $\vx$ as its input.

The performance of these predictors is measured in terms of per-patient prediction accuracy that is determined in the following way. Consider an arbitrary value of a threshold $\tau$. We say that a patient was correctly identified to have septic shock if he or she had a severity score higher than $\tau$ at least once prior to the onset of septic shock.
We say that a patient was falsely identified to have septic shock if he did not experience septic shock during his hospital stay, but his or her severity score rose above $\tau$ at some time point. By considering a series of values of $\tau$, we can obtain receiver operating curve that corresponds to our predictor and the associated area under the curve (AUC). In our experiment, we use the AUC on the validation set of patients to find the optimal values of free parameters of the LR-Shock method.
We report the performance of our classifiers in terms of AUC on the test set (see Table \ref{tab:pred_perf}). This table also contains the predictive performance of the L-DSS and NL-DSS scores.

The learned scores are significantly more accurate than the APACHE II and the SOFA based scores. However, more interestingly, we note that DSS performance is comparable to LR-Shock+x which was trained to optimize predictive performance unlike the DSS scores. Thus, our learning objective addresses the bias introduced due to treatment related confounding without hurting predictive performance in this application. 

\begin{table}[ht]
\centering
\begin{tabular}{l|l|ll}
\hline
& Features & $\LambdaS=\LambdaSOA$ & $\LambdaS = \LambdaAUC$\\
\hline
Predictors & L-DSS &  $0.836$ (0.824-0.849) & $0.853$ (0.841-0.865)\\
based & NL-DSS  & $0.859$ (0.846-0.872)& 0.878 (0.866-0.890)\\
on proposed & L-DSS + Derived&  $0.856$ (0.844-0.868) & $0.857$ (0.845-0.869)\\
scores & NL-DSS + Derived  & $0.861$ (0.849-0.874) & 0.874 (0.862-0.886) \\
\hline\hline
Predictor based &  &  &  \\
on feature vectors & $\vx$ & \multicolumn{2}{|c}{0.864 (0.852-0.875)} \\
alone & & & \\
\hline\hline
Routine & APACHE II & \multicolumn{2}{|c}{0.620 (0.600 - 0.641)}   \\
Clinical & Total SOFA & \multicolumn{2}{|c}{0.602 (0.582-0.622)} \\
Scores & Worst SOFA & \multicolumn{2}{|c}{0.601 (0.581-0.621)} \\
\hline
\end{tabular}
\caption{Per-patient accuracy of early detection of sepsis in terms of the corresponding AUC.
The $95\%$ confidence interval on the AUC is obtained using the bootstrap method and is
given in parentheses.}\label{tab:pred_perf}
\vspace{-25pt}
\end{table}

\section{Related Work on Ranking}

Our work is closely related to the body of literature on pairwise methods for ranking in the field of Information Retrieval (IR). In IR, the ranking problem is typically formulated as the task of sorting retrieved documents 
by their relevance to the query. Pairwise ranking approaches (e.g., \cite{lit:svmrank_joachims_clickthough, lit:zheng2008general}) aim to learn a ranking function that orders pairs of documents in concordance to their relevance. 
These ranking functions are derived using a variety of machine learning techniques. 
\cite{lit:svmrank_joachims_clickthough} proposed a max-margin approach to learning a ranking, dubbed SVMRank. 
A computationally efficient method for training of the SVMRank as a primal-form optimization problem was later proposed by \cite{lit:chapelle_efficient_ranking}. \cite{lit:ranknet} proposed RankNet, a neural network based algorithm for ranking that tunes its parameters to minimize a simple explicit probabilistic cost function that captures the task of pairwise ranking.
In later work, \cite{lit:lambdarank} and \cite{lit:lambda_mart} observed that a ranking function can be learned  
from gradients of the ranking objective function alone, without explicit specification of the whole cost function. 
They proposed two new ranking algorithms that build on this observation, one based on neural networks and one based on gradient boosting trees. \cite{lit:zheng2008general} proposed a general boosting framework for learning ranking functions for a wide family of cost functions. Additional approaches include FRank \citep{lit:tsai2007frank}, nested ranker \citep{lit:matveeva2006high}, and multiple hyperplane ranker \citep{lit:qin2007ranking}.
In some disease domains, as is the case in ours, one might be able to obtain rank data rather than pair-wise comparisons. In these cases, ordinal regression based approaches have been developed for ranking \citep{lit:herbrich1999large,lit:chu2007support}. 
However, acquiring ranked samples from the clinician is often not practical. Moreover, by relying on pariwise comparisons, our framework opens up the possibility 
of exploring new forms of supervision that can be automatically generated. 
For example, two time slices with the same severity grade may be ordered based on their time to an adverse event in the case when no interventions have been administered between these time slices.

\section{Discussion and Future Work}
This paper proposes DSSL, a novel ranking-based framework for scalable and automated learning of disease severity scores in new disease domains and populations.
DSSL only requires a means for obtaining clinical comparisons --- ordered pairs comparing disease severity state at different times.
We argue that this form of supervision is more natural to elicit than asking clinical experts to map the disease severity score, or to encode an accurate model of disease progression.
Moreover, supervision of this type can also be obtained in an automated way by leveraging existing clinical guidelines.

We test DSSL by applying it to a large, real-world electronic health record dataset and to synthetic clinical records. 
Using synthetic clinical records, we show that scores learned using DSSL are less sensitive to changes in treatment administration 
patterns between the train and test environments compared to the regression based approach that is currently used.
Using a large real-world dataset of ICU clinical records, we show that the scores learned using DSSL are significantly more accurate, 
both for severity assessment and early adverse-event detection, compared to widely used clinical severity scores. 
Further, these scores have face validity---their behavior aligns with what is expected clinically. 
They trend upwards leading up to an adverse event, and show decline post-treatment.

DSSL has a number of other advantages. It allows experts to automatically tune the quality of the score by increasing the granularity and amount of supervision given. 
Additionally, the quality of the learned scores can be improved by incorporating additional constraints related to disease progression. 
For example, expected clinical response to therapy can be directly incorporated as a constraint within the optimization objective.

One limitation of our current work is the heavy reliance on the availability of a large number of clinical comparison pairs
\footnote{In other experiments \citep{lit:dss_amia}, we have shown that the performance of our framework degrades gracefully as the number of clinical comparisons is reduced.
In particular, on the task of severity assessment, scores trained on approximately $150$ clinical comparisons are as accurate as APACHE and SOFA.} 
In domains where existing clinical guidelines cannot be leveraged, it is unrealistic to obtain thousands of clinical comparison pairs from experts. 
In these domains, the use of active learning may help mitigate this limitation. Supervision in the form of additional constraints related to disease progression may also prove helpful. Another aspect that deserves further exploration is how the proposed scores can be made interpretable in practice. While NL-DSS yields high performance, the score is constructed using a bag of regression trees. Thus, is it not obvious how one might make the score interpretable at the point of care. In practice, scores are often deployed with a specific use case in mind. For example, if the score were to be used for early detection, the precision-recall curve can be used to identify suitable thresholds for taking action based on the DSS score. Another approach might be to identify and to display which factors led to the increase or decrease in the score value. 
Finally, using simulated data, our experiments show that the scores learned using DSSL are less dependent on the practice patterns of the regime where the model was developed. 
While promising, further analysis is needed to understand susceptibility of DSSL to treatment administration patterns.

In summary, electronic tools that can integrate the diverse and the large set of measurements collected clinically to produce an accurate, real-time severity score can enable clinicians to provide more timely interventions. Further, these scores should be robust to changes in clinical practice patterns as the mere introduction of a decision-support tool can change clinician behavior. This paper introduces a new ranking-based formulation for the problem of learning (predictive) disease severity scores. By leveraging clinical comparisons, a form of supervision that is less susceptible to clinician practice patterns, DSSL provides a promising alternative to existing methods.

\bibliographystyle{spbasic}      
\bibliography{bibliography}

\appendix
\newpage
\section{Features used for DSS learning}\label{app:feat_desc}
In this section we provide the complete list of all features 
provided to L-DSS and NL-DSS methods for DSS learning in experiments in Section \ref{sec:exp_results}. 
These can be divided into three categories: clinical information, measurements of vital signals, 
and results of laboratory analysis.\\
$ $\\
\textbf{Clinical information}: age of the patient; 
whether patient has a pacemaker; whether patient was diagnosed with AIDS; 
whether patient received treatment that compromised his immune system; 
patient's current weight and his weight on admission; 
presence of ICD-9 codes for diabetes, dialysis, chronic renal insufficiency, heart failure, 
or chronic liver disease; whether patient is currently in the cardiac surgery recovery unit; 
presence or absence of hematologic malignancy; jaundice; 
whether a patient was mechanically ventilated;  
presence of metastatic carcinoma.\\
$ $\\
\textbf{Measurements of vital signals:} Glasgow coma score; heart rate; 
Riker Sedation-Agitation Scale; temperature; respiratory rate; 
systolic blood pressure; shock index defined as the ratio of heart rate to systolic blood pressure; 
peripheral capillary oxygen saturation; \\
$ $\\
\textbf{Results of laboratory analysis:} Blood urea nitrogen levels (BUN); 
hematocrit; international normalized ratio (INR); white blood cell count (WBC);
blood pH level as measured by an arterial line;
partial pressure of arterial oxygen ($PaO_2$); 
fraction of inspired oxygen ($FiO_2$); 
ratio of $PaO_2$ to $FiO_2$; 
partial pressure of $CO_2$;
blood lactate measurements; bilirubin; creatinine; 
potassium and sodium levels; 
platelet count; 
hemoglobin; 
total urine output over the past two hours per kg of weight;
partial thromboplastin time; 
arterial $CO_2$ levels;
levels of aspartate aminotransferase.

\end{document}